# Automatic Assessment of Oral Reading Accuracy for Reading Diagnostics

*Bo Molenaar[1], Cristian Tejedor-García[1], Helmer Strik[1], Catia Cucchiarini[1]*

[1]Centre for Language and Speech Technology, Radboud University Nijmegen, Netherlands

bmolenaar1@gmail.com, {cristian.tejedorgarcia, helmer.strik, catia.cucchiarini}@ru.nl

## Abstract

Automatic assessment of reading fluency using automatic speech recognition (ASR) holds great potential for early detection of reading difficulties and subsequent timely intervention. Precise assessment tools are required, especially for languages other than English. In this study, we evaluate six state-of-the-art ASR-based systems for automatically assessing Dutch oral reading accuracy using Kaldi and Whisper. Results show our most successful system reached substantial agreement with human evaluations (MCC = .63). The same system reached the highest correlation between forced decoding confidence scores and word correctness (r = .45). This system's language model (LM) consisted of manual orthographic transcriptions and reading prompts of the test data, which shows that including reading errors in the LM improves assessment performance. We discuss the implications for developing automatic assessment systems and identify possible avenues of future research.

**Index Terms**: reading diagnostics, automatic reading evaluation, automatic speech recognition, oral reading, child speech

## 1. Introduction

Research on the contribution of ASR technology to developing intelligent systems that can support children learning to read in languages other than English is scarce, especially when it comes to studies that have evaluated the usability of the technology under realistic conditions. Important reasons for this are the limited availability of child speech resources for languages other than English, the complexity of developing applications in line with pedagogical requirements and state-of-the-art (SOTA) technology, and the difficulties in obtaining funding for such interdisciplinary projects.

The majority of the studies conducted so far, including those addressing English, have focused on either the development of reading tutors, that is systems that provide feedback while children are reading (e.g. project LISTEN [1, 2]), or the assessment of reading skills at a global level [3, 4]. A possible, innovative contribution of ASR technology could be at the level of reading diagnostics, but, apart from a few exceptions [5, 6], this avenue of research has remained rather unexplored so far. The idea would be that ASR is employed offline to automatically identify reading errors and that the information provided is further analysed to detect patterns in reading errors that might help to identify possible underlying problems or develop more personalised reading trajectories. This level of detail and control is not possible with (third-party) online ASR. In the context of reading diagnostics ASR should do more than recognise words; it should perform an analysis at a more detailed level on the three aspects of reading fluency: Accuracy, speed and prosody. Previous studies on developing ASR to support learning to read in Dutch have not addressed this specific topic [7, 8, 9, 10]. However, considerable progress has been made in ASR in last few years. Recent research has shown that even in online practice ASR technology could be successfully employed to provide feedback to first graders learning to read in Dutch. [11]. Against this background it seems worthwhile to investigate to what extent current ASR systems can be employed to obtain diagnostic measures of reading proficiency. In the present paper we report on a study we conducted to pursue this research goal. The specific research question we addressed is: *To what extent is it possible to automatically assess oral reading accuracy in children learning to read in Dutch?*

To operationalise our RQ we employed two ASR systems, Kaldi and Whisper. Kaldi is a well-established, free, and open-source toolkit for ASR [12]. In contrast, Whisper is a SOTA Python-based ASR system[1]. Despite its novelty, Whisper has already demonstrated promising results in the analysis of speech data, making it a valuable tool for speech-language pathologists and reading tutors alike [13]. By utilising the advanced capabilities of Kaldi and Whisper, reading tutors can more accurately identify reading problems in a student's speech. This information can in turn be used to develop tailored instruction plans to address these problematic areas. Furthermore, the objective data provided by these technologies can help tutors track a student's progress over time and provide evidence-based assessments to support their instruction. Therefore, the integration of ASR into reading tutor programs has the potential to significantly improve the quality and effectiveness of reading intervention programs in schools, ultimately leading to better outcomes for students.

## 2. Methodology

To automatically assess the accuracy of children reading in Dutch, we analysed alignments of reading prompts (PR) with ASR output (AO) and manual orthographic transcriptions (MO) for the utterances produced by each child in a subset of a corpus of child speech [14]. The MOs contain all words uttered by the child, including reading errors. We call the alignment between PR and MO *Reading Errors Manual* (REM), and the alignment between PR and AO *Reading Errors Automatic* (REA). Comparing the REM and REA alignments tells us whether the ASR captures the same reading errors that are present in the MO. This evaluation is key to automatic accuracy assessment: To optimise the degree of actual reading errors captured by the ASR is to optimise automatic accuracy assessment.

The present method consisted of three stages. First, four Kaldi [12] ASR systems with different language models (LMs) were developed to identify the role of each LM. For compari-son, two Whisper ASR systems (with and without prompt hints) were employed alongside the Kaldi systems. Second, hypothesised and referent strings were aligned using ADAPT [15] to measure agreement between human and automatic assessment. Finally, forced decoding (FD) confidence scores were obtained to evaluate Kaldi ASR systems' confidence for each word present in the referent string prompt.

---
[1] https://arxiv.org/abs/2212.04356

### 2.1. Test Data

All analyses were carried out using recordings of native Dutch children's read speech from the JASMIN corpus [14]. This subset of JASMIN contains 1.78 hours of speech for a total of 1455 sentences and 13,180 words. It features 71 children, aged 7-11 years. The corpus has complementary reading prompts and phonemic and orthographic transcriptions for the recordings. In line with the practice in reading instruction in the Netherlands, all speakers read texts that were appropriate for their reading level on the AVI reading scale [16].

Manual orthographic transcriptions were automatically segmented into utterances matching the corresponding prompt and audio file using a Python script[2].

### 2.2. ASR Systems

Four different ASR systems were built using the Kaldi ASR toolkit [12]. Each system had the same acoustic model (AM) but a different language model (LM). In particular, we built nnet3 Time delay neural network and long short-term memory (TDNN+LSTM) chain ASR systems[3] with about 900 hours of mixed media speech by Dutch adults from the largest open-source Dutch speech dataset available, the Corpus Spoken Dutch [17, 18] (CGN). The acoustic model contained 7 layers, including a 40-dimension bottleneck layer at the 6th layer. The input features were 40-dimension HR MFCCs. Frame labels for TDNN model training were obtained by forced alignment using a GMM-HMM model trained beforehand. The numbers of modeled phones and triphone HMM states in the model were 81 and 3361, respectively. The LMs were 4-grams in ARPA format, built as follows. ASR Kaldi-CGN used the large general-purpose LM from CGN (150k words in the lexicon). ASR Kaldi-PR used an LM containing only words from the reading prompts (PR) from the JASMIN test data. ASR Kaldi-MO used an LM containing only words from the MO from the test data, and ASR Kaldi-PM used an LM containing both PR and MO words. In all cases interpolation was applied to the LM.

In addition, two Whisper-based ASR systems were tested on this task. Whisper ASR models have been recently introduced by OpenAI. They are general-purpose and multi-task, trained in a fully supervised manner, using up to 680k hours of labeled speech data from multiple sources. The models are based on an encoder-decoder Transformer, which is fed by 80-channel log-Mel spectrograms. The encoder is formed by two convolution layers with a kernel size of 3, followed by a sinusoidal positional encoding, and a stacked set of Transformer blocks. The decoder uses the learned positional embeddings and the same number of Transformer blocks from the encoder. For this study, we used the "Whisper large-v2" model, which consists of 1550 million parameter distributed in 32 layers and 20 attention heads. The model is available via Huggingface[4]. Whisper offers one way to suggest vocabulary (clues) to the model in order to increase probabilities of the provided words. Thus, two alternatives were evaluated, without (Whisper-Lv2) and with prompts as clues (Whisper-PR). The decoding was performed using a beam search strategy with 5 beams, an array of temperature weights of [0.2,0.4,0.6,0.8,1].

The computer configuration on which the experiment was conducted had Ubuntu 18.04.1 LTS (64-bit operating system), AMD EPYC 7502P 32-Core (64 threads) processor with 2.5-3.35 GHz, 251GB of RAM and three NVIDIA Tesla T4. On this machine, we trained the models for Kaldi and decoded the test audio files using both Kaldi and Whisper ASR systems.

As standard measures of decoding performance for each ASR system, word and sentence error rates (WER and SER) were calculated. To obtain error rates, ASR output was scored against both prompts and manual orthographic transcriptions using sclite from the SCTK toolkit[5].

### 2.3. Alignment with ADAPT

The performance of the six ASR systems in assessing oral reading accuracy was evaluated using ADAPT, a dynamic programming algorithm [15]. ADAPT takes a referent and hypothesis string for a given sentence, aligns them, and scores the edit distance between them on the word level. Since orthographic transcriptions were used, edit distance was calculated as the distance between two grapheme strings. If a reference word is present in the hypothesis, ADAPT judges it as correct. In this case, the referent string is the reading prompt and the hypothesis is the transcription of the child's utterance.

We adopted Bai et al.'s [19] method of alignment, which takes into account that children's oral reading may contain restarts. In line with how teachers judge such utterances, the algorithm judges a word as read correctly if it finds a complete instance of that word in the whole utterance. This is done by starting matching at the end of the utterance and working backwards, thus avoiding scoring any restarts or partial words before the correct word is found.

| | |
|---|---|
| **PR** | wanneer ze klaar is klappen we opnieuw |
| **MO** | wanneer uh ze k klaar is klap klappen we opnieuw |
| **Aligned PR** | wanneer ---ze --klaar is -----klappen we opnieuw |
| **Aligned MO** | wanneer uh\|ze k\|klaar is klap\|klappen we opnieuw |

Figure 1: *Example of ADAPT alignment of a prompt and manual orthographic transcription.*

An example from the test data of PR, MO, and their alignment can be found in Figure 1. Hyphens mark insertions (characters present in the hypothesis but not in the reference) or deletions (vice versa). Vertical lines mark word boundaries. In this example, ADAPT calculates a distance score of 10 insertions between PR and MO, but marks all words as correct as indeed all PR words are present in the MO.

The baseline alignment of PR and MO constitutes Reading Errors Manual (REM). In addition, the output from each ASR (AO) was aligned with prompts to generate Reading Errors Automatic (REA). REA for each ASR were then evaluated using the correctness scores provided by the ADAPT as follows. If we consider the correctness judgements from REM and REA as two different ratings, by a human and automatic rater, we can calculate inter-rater agreement on these scores. Agreement was calculated using several metrics. Confusion matrices were generated for REM and each REA in R Statistical Software[20].

---

[2]https://github.com/bomolenaar/jasmin_data_prep
[3]https://github.com/cristiantg/kaldi_egs_CGN/tree/onPonyLand
[4]https://huggingface.co/openai/whisper-large
[5]https://github.com/usnistgov/SCTK/blob/master/doc/sclite.htm

These were then used to calculate precision, recall, specificity, F1 and Matthews correlation coefficient (MCC) [21, 22]. Since our data is very skewed (most words are read correctly), we report MCC here instead of the more common Cohen's kappa [23]. Kappa assumes equal distribution over classes and MCC does not, making it a better fit to our data [22].

### 2.4. Forced Decoding

The automatic accuracy assessment of each Kaldi ASR system was additionally evaluated using Forced Decoding (FD)[6]. Forced decoding is an ASR technique that forces a speech signal to be decoded to a target string. It compares the speech signal to the AM for the target and then provides posterior probability scores as a measure of confidence that the target AM matched the provided utterance. These confidence scores are generated for each word and for the whole utterance (mean of the word-level scores). This is possible thanks to the representation of the alternative word-sequences that are "sufficiently likely" for a particular utterance called lattices in the Kaldi argot. Some confidence scores were erroneously calculated as either below 0 or larger than 1, which is outside the domain of probability. These scores were rounded to 0 and 1, respectively.

FD confidence scores were obtained for Kaldi ASR systems. Note that the AM for each Kaldi system was the same, which means differences between FD scores for the same word are based on the lattices obtained from the combination of the general-purpose AM and the specific LM of each Kaldi system.

Forced decoding was done for each Kaldi ASR that contained words from the reading prompt in the lexicon, namely -CGN, -PR and -PM. The reason for this is the fact that FD requires all target words to be in the LM of the ASR. Carrying out FD for Kaldi-MO would result in discrepancies between the ADAPT alignment and FD output, since its LM consisted only of manual orthographic transcriptions. Furthermore, since Kaldi-PM has an LM containing both MO and PR, FD was performed only with the prompt as the decoding target.

## 3. Results

### 3.1. Baseline and ASR Error Rates

In order to assess to what extent our dataset can be assessed automatically, it should first become clear how many reading inaccuracies we are trying to capture from the speech data. In addition, we need generic statistics on the accuracy of each ASR system. The baseline error rate and error rates for each ASR are reported here.

Table 1: *Baseline ADAPT alignment of PR and MO.*

| Item | N | N correct | Error rate (%) | ADAPT dist. score Mean (SD) |
|---|---|---|---|---|
| Words | 13,149 | 12,778 | 2.82 | 2.98 (2.72) |
| Sentences | 1,455 | 1,178 | 19.04 | 6.42 (7.57) |

Table 1 shows descriptive statistics for the baseline alignment of MO and PR for each utterance. Only 2.82% of words were read incorrectly, which suggests that most children in the corpus were proficient readers at the level of the text they read, as is normal for Dutch pupils. The mean ADAPT distance score, i.e. the mean edit distance between target and actual utterance was 2.98 for incorrect words and 6.42 for sentences. Notably, the standard deviation was larger than the mean for sentences, indicating some large outliers. Outliers with edit distance ≥ 20 were excluded from further analysis.

[6]https://github.com/homink/kaldi-asr.forced_decoding

WER, SER and accuracy for all ASR systems are reported in Table 2 for both ground truth conditions (MO and PR). The best WER (4.5%), SER (14.8%) and accuracy (99.0%) values for ASR output compared to PR was reached by Kaldi-PR. Conversely, Kaldi-MO reached best WER (5.0%), SER (26.8%) and accuracy (95.5%) for AO compared to MO. Whisper's general-purpose model reached better WER, SER and accuracy than Kaldi-CGN in both ground truth conditions. ASR Kaldi-PM's error rates were in between those of Kaldi-PR and Kaldi-MO in both ground truth cases. Notably, while the generic Whisper-Lv2 outperformed generic Kaldi-CGN, the specific-LM Kaldi systems performed much better than specific-LM Whisper.

Table 2: *Error rates versus PR and MO for each ASR.*

| ASR system | LM | Ground truth | WER (%) | SER (%) | ACC (%) |
|---|---|---|---|---|---|
| Kaldi-CGN | CGN | MO | 29.6 | 84.0 | 73.3 |
| | | PR | 33.6 | 86.0 | 74.8 |
| Kaldi-PR | PR | MO | 8.3 | 41.1 | 92.4 |
| | | PR | **4.5** | **14.8** | **99.0** |
| Kaldi-MO | MO | MO | **5.0** | **26.8** | **95.5** |
| | | PR | 8.7 | 33.9 | 97.1 |
| Kaldi-PM | PR + MO | MO | 5.2 | 28.5 | 95.2 |
| | | PR | 7.6 | 28.3 | 97.9 |
| Whisper-PR | PR + large_v2 | MO | 15.5 | 41.7 | 87.7 |
| | | PR | 9.8 | 22.3 | 94.5 |
| Whisper-Lv2 | large_v2 | MO | 13.4 | 53.0 | 87.2 |
| | | PR | 10.9 | 47.6 | 91.1 |

### 3.2. Agreement Metrics

Metrics on the comparison of REA with REM are reported in Table 3. ASR Kaldi-CGN reached highest precision (.994) and specificity (.842), but much lower recall (.773) than Kaldi-PR, -MO and -PM (.990 – .995), which had LMs customised to the test data. Only Kaldi-CGN reached a level of specificity somewhat close to its recall, which shows that this system had fewer false positives and thus was better at identifying true negatives.

Table 3: *Agreement metrics between REM and REA.*

| ASR system | Precision | Recall | F1 | Specificity | MCC |
|---|---|---|---|---|---|
| Kaldi-CGN | **.994** | .773 | .870 | **.842** | .24 |
| Kaldi-PR | .977 | **.995** | .985 | .182 | .28 |
| Kaldi-MO | .990 | .990 | .990 | .655 | **.65** |
| Kaldi-PM | .987 | .994 | **.991** | .554 | .63 |
| Whisper-PR | .983 | .957 | .970 | .420 | .28 |
| Whisper-Lv2 | .972 | .946 | .959 | .067 | .01 |

In the same vein, Kaldi-PR and Kaldi-PM have very high F1 scores (.985 and .991) but low specificity (.182 and .554), indicative of few missed true positives and many missed true negatives. This can be clearly seen in the confusion matrix for ASR Kaldi-PM in Table 4. The ASR output is heavily skewed towards acceptances, with only 2.12% of words judged as read incorrectly. This is close to the actual error rate in this dataset: Only 2.82% of words are read incorrectly. This similar error rate in REM and REA suggests the ASR performs well at spotting reading errors.

MCC was calculated to define agreement between REM and REA. The highest MCC (.65) was found for REA by ASR Kaldi-MO, closely followed by Kaldi-PM (.62). Both indicate substantial agreement between REM and REA. REA by other systems was much less accurate. Kaldi-CGN and -PR had an MCC of .24 and .28 (fair) with REM, respectively. Whisper-PR and Kaldi-PR reached equal agreement with REM at MCC=.28 (fair), whereas Whisper-Lv2 had no agreement with REM at MCC=.01.

Table 4: *Confusion matrix of word correctness in PR:MO alignment (REM) and PR:AO alignment (REA) for Kaldi-PM.*

|  |  | REM Incorrect | REM Correct |
|---|---|---|---|
| REA | Incorrect | 205 (1.56%) | 72 (0.55%) |
|  | Correct | 166 (1.26%) | 12,706 (96.63%) |

### 3.3. Correlations between FD and ADAPT

We expected a relationship between FD confidence scores and ADAPT correctness judgements, since FD confidence scores should be lower for incorrect words and higher for correct ones. We therefore calculated point-biserial correlations between FD confidence scores and ADAPT correctness for the Kaldi ASR systems that contained prompt words: Kaldi-CGN, Kaldi-PR and Kaldi-PM. We also calculated Pearson correlations between FD confidence scores and ADAPT distance scores for all words. These correlations are presented in Table 5. Because Whisper was outperformed by Kaldi in error rate and agreement, we did not not calculate comparable correlations between Whisper's confidence scores and ADAPT correctness.

All correlations between FD confidence scores and ADAPT correctness judgements (A.Cor) were significant at p<.001. The highest correlation was found for Kaldi-PM at $r$ = .45. Much lower correlations were found for Kaldi-CGN ($r$ = .15) and Kaldi-PR ($r$ = .18). Correlations between FD confidence scores and ADAPT distance scores (A.Dist) were small. The highest correlation was found for Kaldi-PM at $r$ = -.20.

## 4. Discussion

Our findings show that ASR for children's oral reading with a narrow LM may lead to better accuracy in word recognition, but is not so good at identifying reading errors. This applies to both Kaldi and Whisper: Kaldi-PR and Whisper-PR both had better WER, SER and accuracy but not better agreement than the generic models Kaldi-CGN and Whisper-Lv2. Furthermore, Kaldi outperformed Whisper even despite the use of reading prompts as clues: Kaldi-MO and Kaldi-PM both had better error rates and agreement than Whisper-Lv2 and Whisper-PR. The highest agreement was found for the ASR system with a language model consisting of MO of the target speech (ASR Kaldi-MO, MCC = .65). This shows that prior knowledge of the words that the child may use helps the ASR identify reading errors. However, while better performance is good, this is not necessarily a desirable result as the ASR should be able to assess reading accuracy automatically and without the support of human transcribers.

The FD confidence scores for ASR Kaldi-PM reached the highest correlation with both REM ($r$ = .45) and ADAPT distance scores ($r$ = -.20), indicating that Kaldi-PM outperforms Kaldi-MO because it scores well on all metrics: Error rate, agreement, and correlation of confidence scores. Since Kaldi-PM is trained with generic AM but a specific LM including both PR and MO, it was expected to be most sensitive to actual reading inaccuracies.

Table 5: *Correlations of FD confidence scores with ADAPT correctness (A.Cor) and with ADAPT distance scores (A.Dist).*

| ASR system | LM | A.Cor | 95% CI | A.Dist |
|---|---|---|---|---|
| Kaldi-CGN | CGN | .15*** | .13 – .64 | -.08*** |
| Kaldi-PR | PR | .18*** | .17 – .20 | -.08*** |
| Kaldi-PM | PR + MO | **.45*** | .44 – .46 | **-.20*** |

More surprising is the low correlation ($r$ = .18) between Kaldi-PR FD confidence scores and REM, since Kaldi-PR reached the best WER, SER and accuracy on our data with PR as ground truth. This shows that a low error rate does not necessarily correspond to accurately identifying reading inaccuracies. Moreover, there is only a minor difference between FD correlations for Kaldi-CGN and Kaldi-PR, but a much larger difference to Kaldi-PM. This shows that when the AM is equal, including MO in the LM improves the ASR's confidence the most. To answer our research question, it is important to evaluate the quality of the six systems in the right terms, as the evaluation should focus on the output of the ASR systems. It has become clear that they do not reach a high level of agreement with human transcribers when it comes to detecting reading errors. However, by combining several metrics, we can see that Kaldi-PM assessed over 98% of words correctly while reaching substantial agreement and significant but moderate correlation with REM (MCC = .63; $r$ = .45). This combination of metrics seems to be promising for automatic assessment. While higher sensitivity is desirable, the system can be useful even without optimal metrics. For instance, consider the context in which automated oral reading assessment is employed: Children practising their reading - say, two-syllable nouns in grade 1. This context can be utilised to fine-tune the ASR's LM, which in turn improves performance. Finally, since we focus on offline assessment, false positives and false negatives are not as detrimental as they would be in an online feedback setting. Assessment can be done over several measurements, which means the system will have more data to work with and thus be more robust in modelling reading errors.

We identified two key areas to improve our method. Firstly, this research showed that using an LM containing PR and MO results in highest agreement with human assessment of reading accuracy. This means more work needs to be done to improve performance without MOs. Secondly, much can be gained by using ASR systems trained on child speech. Future research may address these shortcomings.

## 5. Conclusion

We evaluated six ASR systems for automatic assessment of Dutch children's oral reading accuracy. The best performing system, Kaldi-PM, used an LM including both PR and MO. It reached substantial agreement and a moderate correlation with human assessment (MCC = .63; $r$ = .45). Though its performance was not perfect, we posit that this system can provide a useful basis for automatic assessment of oral reading accuracy.

## 6. Acknowledgements

This work is part of the research program Open Competition in the Humanities and Social Sciences with project number 406.20.TW.009, which is financed by the Netherlands Organisation for Scientific Research (NWO).